\documentclass{article}
\usepackage{spconf,amsmath,graphicx,hyperref}
\usepackage{amssymb,amsfonts}
\usepackage{algorithmic}
\usepackage{dblfloatfix}
\usepackage{textcomp}
\usepackage{array}
\usepackage{multirow}
\usepackage{makecell}
\usepackage{cite}
\usepackage{booktabs}
\usepackage[mathscr]{eucal}
\usepackage{relsize}
\usepackage{xcolor}
\usepackage[table]{xcolor}

\title{UNCERTAINTY-BASED ENSEMBLE LEARNING IN CMR SEMANTIC SEGMENTATION}
\name{Yiwei Liu$^{1,2}$ \quad Liang Zhong$^{2}$ \quad Lingyi Wen$^{3}$ \quad Yuankai Wu$^{1,\text{\dag}}$
\thanks{$^{\text{\dag}}$Corresponding Author, e-mail: wuyk0@scu.edu.cn. ICASSP 2026.}
}
\address{
$^1$Sichuan University \quad
$^2$National University of Singapore \quad
$^3$West China Women's and Children's Hospital \quad}

\begin{document}
%
\maketitle
\begin{abstract}
Existing methods derive clinical functional metrics from ventricular semantic segmentation in cardiac cine sequences. While performing well on overall segmentation, they struggle with the end slices. To address this, we extract global uncertainty from segmentation variance and use it in our ensemble learning method, Streaming, for classifier weighting, balancing overall and end-slice performance. We introduce the End Coefficient (EC) to quantify end-slice accuracy. Experiments on ACDC and M\&Ms datasets show that our framework achieves near state-of-the-art Dice Similarity Coefficient (DSC) and outperforms all models on end-slice performance, improving patient-specific segmentation accuracy. We open-sourced our code on \url{https://github.com/LEw1sin/Uncertainty-Ensemble}.
\end{abstract}
\section{Introduction}
\label{sec:introduction}
The cardiac cine technique in cardiac magnetic resonance (CMR) imaging has become the gold standard for the non-invasive assessment of various cardiovascular functions \cite{ismail2022role}. For ventricular segmentation of the left ventricle (LV), right ventricle (RV), and left ventricle myocardium (MYO) from the background (BG), which is a pixel-wise semantic segmentation task, the rise of deep learning has freed cardiologists from labor-intensive manual labeling in recent years. 

Cardiac cine sequences are 4D, with three spatial dimensions (depth × height × width, 3D) and time. Clinically, emphasis is placed on end-diastolic (ED) and end-systolic (ES) phases, so the raw data mainly consists of 3D frames from these two phases. Researchers address this task through two approaches. One treats the entire 3D frame as input and employs a 3D UNet for segmentation (3D-based) \cite{isensee2018automatic,patravali20182d}. However, several studies pointed out that sequential convolution and pooling operations during the UNet encoder phase inherently struggle to learn long-range relationships between pixels \cite{ruan2024vm}, sometimes leading to misidentification of multiple ventricles. The other approach slices the 3D frame into individual 2D slices \cite{chen2021transunet,cao2022swin,sun2020saunet,rahman2024multi,tragakis2023fully}. It enhances segmentation accuracy on a single 2D slice by incorporating mechanisms such as Vision Transformer (attention-based), which can achieve strong numerical performance on overall segmentation. However, this approach significantly underperforms on end slices, e.g., the first and last slice of the 3D frame, where ventricular volumes are more pronounced compared to the middle. The upper end slices correspond to the basal segment of the heart, while the lower ones correspond to the apical segment. Misidentification of LV during individual segmentation can lead to incorrect calculation of indices such as the ejection fraction (EF) \cite{bhan2022deep}. Moreover, minor differences in segmentation performance metrics may not be discernible upon visual inspection, nor do they necessarily impact clinical decision-making \cite{bhan2022deep,penso2022cardiovascular}. Therefore, there is still a gap in maintaining high segmentation accuracy overall and on the end slices.

For each 2D slice (height × width) that maintains continuity across spatial (depth) and temporal dimensions, we argue that the above dilemma could be solved by ensemble learning leveraging this intrinsic characteristic of cardiac cine, which is that high segmentation accuracy in middle slices can provide priors for end slices, thereby improving the segmentation performance of end slices. This raises two issues to discuss. \textbf{Q1: How to represent such spatial continuity? Q2: How to apply such continuity to ensemble learning?}

For Q1, Kendall et al. \cite{kendall2017uncertainties} proposed estimating model uncertainty in semantic segmentation by calculating the variance of pixel-wise probabilities. They observed that models exhibit high uncertainty at class boundaries. Given the similarity of ventricular boundaries across adjacent 2D slices due to continuity, the ventricular positions in middle slices can serve as references for segmenting end slices. Therefore, we use uncertainty to represent the spatial continuity.

\begin{figure*}
    \centering
    \includegraphics[width=0.85\linewidth]{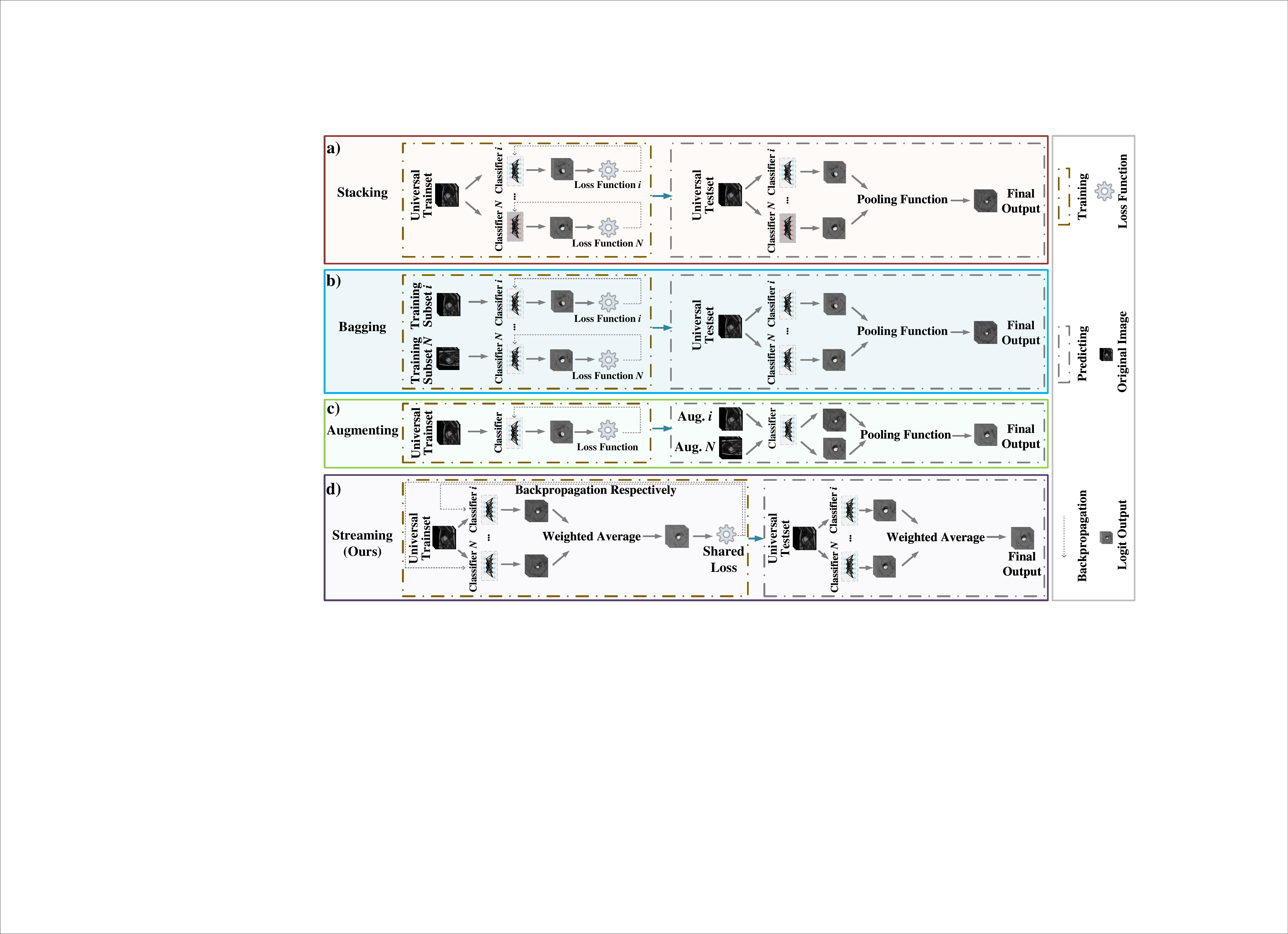}
    \caption{\textbf{Visual differences between traditional ensemble learning and our Streaming}. a) Classifiers in different colors denote different model types. b) Classifiers in identical colors represent identical model types. c) Aug. abbreviates testsets augmented via methods (e.g., rotation). d) Shared Loss indicates all classifiers undergo joint training with a shared loss function. For other legends, Loss Function $i$ signifies classifiers trained with distinct losses, Universal Set means all classifiers use a shared, identical dataset, brown and gray boxes correspond to the training and predicting stage, respectively.}
    \label{fig:end-to-end}
\end{figure*}

For Q2, ensemble learning involves training multiple base learners on the same or different data subsets and then aggregating (i.e., pooling) them according to a certain policy, such as Voting. Base learners can be machine learning models like decision trees, and deep learning models like CNNs have also begun to be used recently. Müller et al. \cite{muller2022analysis} reviewed three ensemble techniques in medical imaging: \textit{Bagging}, \textit{Stacking}, and \textit{Augmenting}, whose overviews are shown in Fig.~\ref{fig:end-to-end}. \textit{Stacking} trains different networks (i.e., heterogeneous) on the same dataset and pools their outputs, while \textit{Bagging} trains the same type of network (i.e., homogeneous) on different trainsets and aggregates predictions on a shared testset. \textit{Augmenting}, in contrast, uses a single network to process augmented test data, combining outputs via pooling. These methods do not fully exploit deep learning's end-to-end (E2E) nature. 

In summary, our contributions are as follows:

\textbf{Quantification on End Slices}: We propose a metric in (\ref{eq:EC}) to evaluate segmentation performance on the end slices and conduct benchmarks on current cutting-edge models.

\textbf{Ensemble Strategies}: We design an E2E pipeline, Streaming, which uses neural networks as base classifiers.

\textbf{Uncertainty Mechanism}: We introduce the uncertainty which is derived from the spatial continuity of 3D frames for weight allocation in our Streaming. We achieve near state-of-the-art (SOTA) performance in overall 3D frame segmentation and significantly improve accuracy on end slices.

\section{Methodology}
\subsection{End-to-end Ensemble Strategy}
\label{sec:end-to-end training}
We design a framework shown in Fig. \ref{fig:end-to-end} d), Steaming, which is in an E2E manner both in the training and predicting stages and unifies the dataset and loss function across all components, streamlining training. 
For the \(i\)-th classifier, the gradient for training is:
\begin{equation}
\frac{\partial{\mathcal{L}}}{\partial{\Theta_{i}}} = \frac{\partial{\mathcal{L}}}{\partial{\hat{y}}} \cdot 
\frac{\partial{\hat{y}}}{\partial{\hat{y}_{i}}} \cdot \frac{\partial \hat{y}_{i}}{\partial \Theta_{i}},
\end{equation}
where $\mathcal{L}$ is the loss function, \(\hat{y}\) is the ensemble output, and \(\hat{y}_{i}\) is the output of the \(i\)-th classifier with \(\Theta_{i}\) as its parameters. If each classifier’s gradient is differentiable, they can share the loss function for backpropagation via the chain rule.
The joint training strategy resembles Mao et al.'s E2E paradigm \cite{mao2019end}, except they dynamically modulate weights via sparse constraints at the classifier level, whereas our approach directly optimizes parameters through pixel-wise weighted loss.

\subsection{Uncertainty Mechanism} 

To synthesize the outputs from multiple classifiers, an intuition is to construct a convex combination, and then use the Softmax function to implement Majority Vote Soft pooling:
\begin{equation}
\hat{y}(\textbf{x}) = \text{softmax}\left(\sum_{\textit{i}=1}^{\textit{N}} \omega_{\textit{i}} \hat{y}_{\textit{i}}(\textbf{x})\right), 
\\ \text{s.t.}\sum_{\textit{i}=1}^{\textit{N}} \omega_{\textit{i}} = 1, \omega_{\textit{i}} \geq 0,
\end{equation}
where $\textbf{x}$ is an 3D frame from the 4D cine sample space $\mathcal{X}$, \(\hat{y}_{i}(\textbf{x})\) is the probabilities of the \textit{i}-th classifier’s segmentation, \( N \) is the number of classifiers, $\omega_{i}$ is the prior fixed weight (say 0.5) of the $i$-th classifier. We adopted this approach as a baseline for ablation in our experiments (denoted as \textit{Fixed}). 

A more refined method is dynamically allocating weights at the pixel-wise level. Therefore, we propose:
\begin{equation}
\hat{y}(\textbf{x}) = \text{softmax}\left(\sum_{\textit{i}=1}^{\textit{N}} \bar{\boldsymbol{\omega}}_{i} \odot \hat{y}_{\textit{i}}(\textbf{x})\right),
\end{equation}
\begin{equation*}
\text{s.t.}\left\{
\begin{aligned}
& \boldsymbol{\mu}_{i} = Mean_{D}(\hat{y}_{\textit{i}}(\textbf{x})), \quad \boldsymbol{\mu}_{i} \in \mathbb{R}^{+}_{4\times H \times W}\\
& \boldsymbol{\sigma}_{i} = Var_{channel}(\boldsymbol{\mu}_{i}), \quad \boldsymbol{\sigma}_{i} \in  \mathbb{R}^{+}_{H \times W}\\
& \bar{\boldsymbol{\omega}}_{i} = \frac{\exp{(\boldsymbol{\sigma}_{i}})}{\sum_{\textit{i}=1}^{\textit{N}}\exp{(\boldsymbol{\sigma}_{i}})}, \quad \sum_{\textit{i}=1}^{\textit{N}}\bar{\boldsymbol{\omega}}_{i} = J_{H \times W}\\
\end{aligned}
\right.
\end{equation*}
where the mean \(\boldsymbol{\mu}_{i}\) is computed along the depth dimension, representing the likelihood of each channel at each pixel. $4$ represents the four channels of BG, RV, MYO, and LV, while \(D, H, W\) denote the depth, height, and width of the 3D frame. The uncertainty of the \(i\)-th classifier is the variance \(\boldsymbol{\sigma}_{i}\) of \(\boldsymbol{\mu}_{i}\) along the \(\textit{channel}\) dimension, where higher variance indicates higher consistency to a channel across multiple slices, thus lower uncertainty (i.e., higher variance between channels indicates the \(i\)-th classifier's high confidence in a specific class, allowed to dominate the ensemble). \( J \) is an all-ones matrix, indicating a finer-grained convex combination. $\bar{\boldsymbol{\omega}}_{i}$, positively correlated with $\boldsymbol{\sigma}_{i}$, is obtained through regularization, then broadcasted to perform the Hadamard product $\odot$ with $\hat{y}_{i}(\textbf{x})$. Thus, the forward process uses uncertainty by having the \(i\)-th classifier independently predict each 3D frame to store \(\boldsymbol{\sigma}_{i}\). During pooling, it dynamically assigns pixel-wise weights $\bar{\boldsymbol{\omega}}_{i}$, which leverages spatial continuity to make slices could refer to each other and stabilize performance on end slices.

\subsection{Loss Function}
Since RV, MYO, and LV in CMR images exhibit a highly imbalanced distribution, with BG occupying most of the image, our loss function is designed as follows: 
\begin{equation}
    \left\{
\begin{aligned}
& \mathcal{L}_{\text{Dice}}(\hat{y}, y) = \sum_{j=1}^{4}\lambda_{j}(1-\text{DSC}(\hat{y_{j}},y_{j})) \\
& \mathcal{L}_{\text{Focal}}(\hat{y}, y) = -\sum_{j=1}^{4}\alpha_{j}(1-\hat{y_{j}})^{\gamma}y_{j}\log{\hat{y_{j}}} \\
& \mathcal{L}_{\text{Total}} = \mathcal{L}_{\text{Dice} } + 10\mathcal{L}_{\text{Focal}}
\end{aligned}
\right.
\label{eq:loss function}
\end{equation}
where \(\hat{y}_{j}\) and \(y_{j}\) are the predicted probabilities and ground truth 0-1 mask for each class \(j\). \(\mathcal{L}_{\text{Dice}}\) is the weighted Dice loss, with DSC from (\ref{eq:DSC}) and weights \(\lambda_{j} = 2\) for RV, \(1\) for BG, MYO, and LV. \(\mathcal{L}_{\text{Focal}}\) is the Focal Loss, with \(\alpha_{j} = 0.1\) and \(\gamma = 2\). The total loss \(\mathcal{L}_{\text{Total}}\) is a weighted sum of Dice and Focal losses with weights \(1\) and \(10\), respectively.

\begin{figure}[t]
    \centering
    \includegraphics[width=1.0\linewidth]{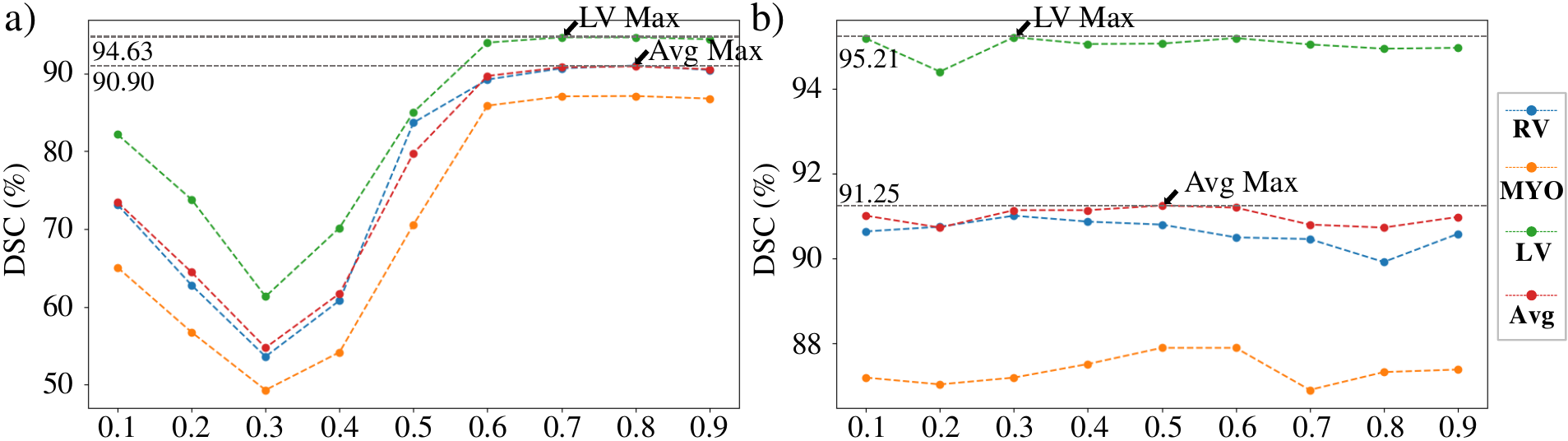}
    \caption{a) shows results from 1UNet + 1D3P, with the x-axis varying UNet's weight. b) shows results from 2UNet, with the x-axis varying UNet 1's weight.}
    \label{fig:convex comb comprison viz}
\end{figure}

\begin{figure}[t]
    \centering
    \includegraphics[width=1.0\linewidth]{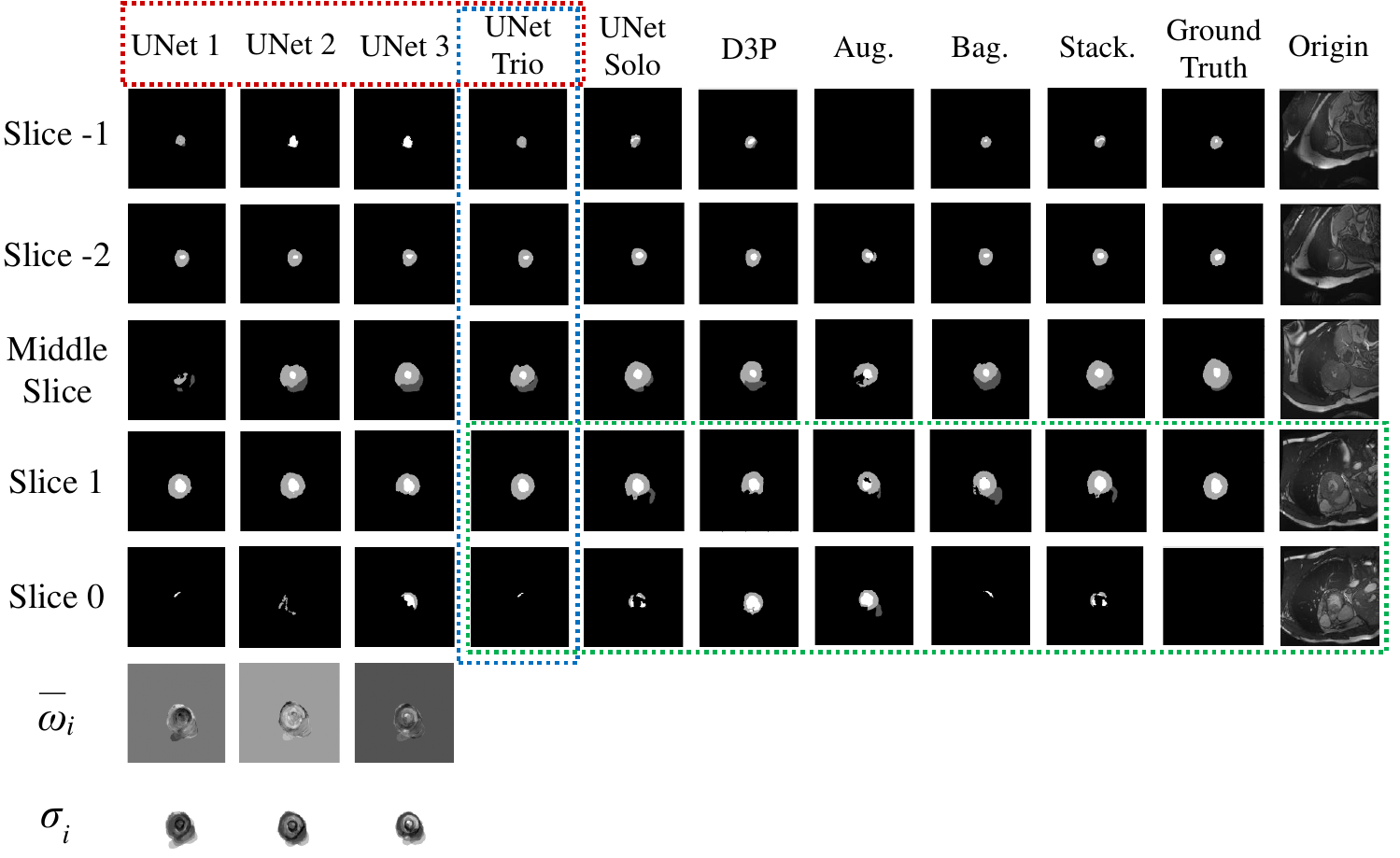}
    \caption{UNet Trio is 3UNet (Uncertainty), and UNet \textit{i} is one of its components. Solo means working individually. $-1$ and $-2$ are the last two slices, $0$ and $1$ the first two.}
    \label{fig:cmr viz}
\end{figure}

\section{Experiments}
\subsection{Setting}
\textbf{\textit{1) Datasets}}.
The first dataset we used is ACDC dataset \cite{bernard2018deep}, which includes 150 patient cases. The second is M\&Ms Challenge \cite{campello2021multi}, consisting of 375 participants from six hospitals. Expert clinicians manually annotated both datasets' LV, RV, and MYO at the ED and ES phases. The trainset and testset have been pre-split by the committee.
\textbf{\textit{2) Classifier Selection}}.
We chose UNet and DeepLabV3+ (denoted as D3P) as our base classifiers on an NVIDIA RTX A5000 with 24~GB of GPU memory. Both have a maximum of 256 channels in the bottleneck. The batch size was set to 8, and the epochs to 500. All classifiers shared a single optimizer \textit{RMSprop} with a learning rate of $1 \times 10^{-4}$, a weight decay of $1 \times 10^{-7}$, and a momentum of $0.9$.
\textbf{\textit{3) Metrics}}.
\label{sec:metrics}
We employed two metrics:
\begin{itemize}
    \item Dice Similarity Coefficient (DSC):
    \begin{equation}
        DSC=\frac{2 \vert \hat{y}_{j} \cap y_{j} \vert}{\vert \hat{y}_{j} \vert + \vert y_{j} \vert},
    \label{eq:DSC}
    \end{equation}
    where \(\hat{y}_{j}\) and \(y_{j}\) are the predicted probabilities and ground truth 0-1 mask for each class \(j\).
    \item End Coefficient (EC):
    \begin{equation}
    EC=\frac{1}{K}\sum_{k=1}^{K} {\scalebox{1.0}{$\mathbb{I}$}(DSC_{Avg}(\hat{y}_{end}, y_{end}) > 0.8)}
    \label{eq:EC}
    \end{equation}
    where \(\hat{y}_{end}\) is the segmented mask of the first two and last two slices of a 3D frame, and \(y_{end}\) is the ground truth. $\mathbb{I}$ is the indicator function, which takes the value $1$ if the Average DSC on end slices is larger than the threshold $0.8$, and $0$ otherwise. $K$ is the number of 3D frames. Because annotation protocols of MYO tissue may yield incomplete apical labels, EC calculation considers only fully annotated end slices.
\end{itemize}

\subsection{Results}
\label{sec: exp results}

\textbf{\textit{Comparison on ACDC and M\&Ms}}. We conducted comparisons and benchmarks on ACDC and M\&Ms datasets, with results presented in the first and second major row in Tab.~\ref{tab:results} respectively. In the first major row, No. 1-2 are 3D-based, while No. 3-6 are attention-based. In the second major row, all baselines are variants of UNet. The results demonstrate that our framework achieves near SOTA performance on both ACDC and M\&Ms in terms of average DSC. Moreover, it outperforms existing works on EC, making it more suitable for clinical segmentation of patient-specific samples due to the correct identification of the apex and base of the LV.

\begin{figure}[t]
    \centering
    \includegraphics[width=1.0\linewidth]{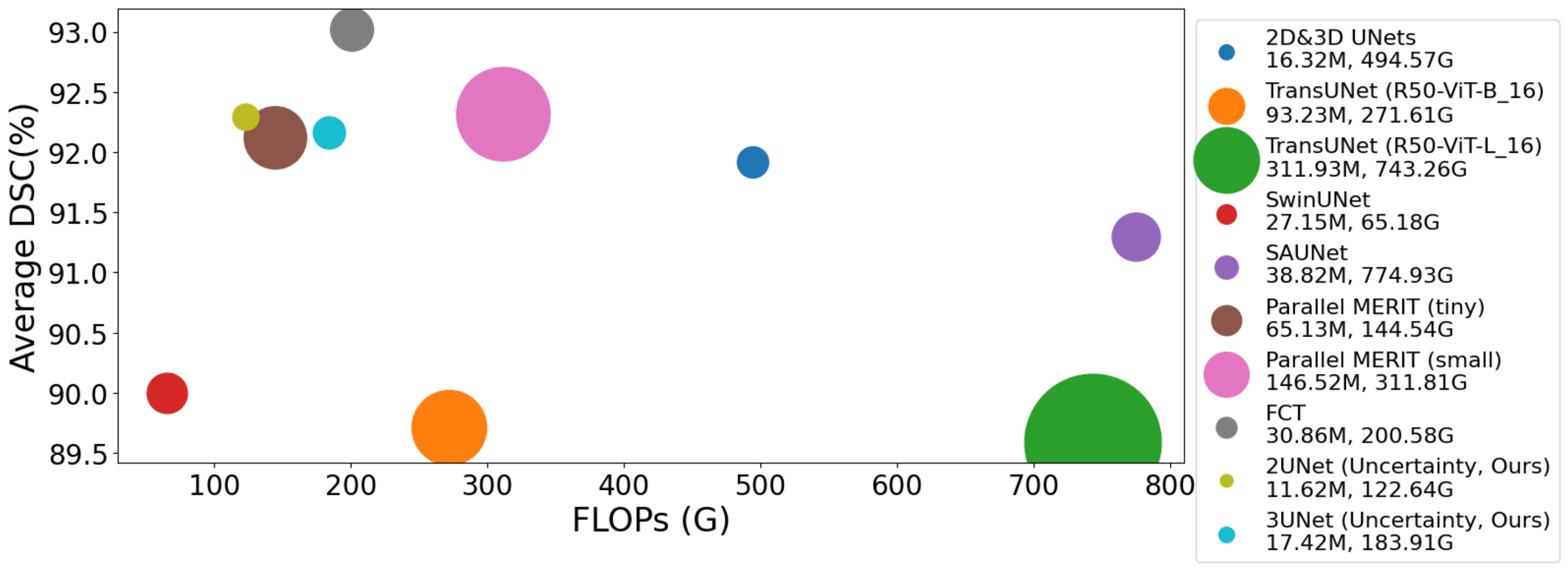}
    \caption{Comparison of model parameters, FLOPs, and Average DSC. Marker size shows parameter count.}
    \label{fig:param}
\end{figure}

\begin{table}[t]
\caption{Experimental Results on ACDC and M\&Ms ($\uparrow$)}
\small
\centering
\setlength\tabcolsep{1.2pt}
\begin{tabular}{ccccccc}
\toprule
No. & Architectures & Average & RV & MYO  & LV & EC\\ 
\midrule
\multicolumn{7}{c}{\cellcolor{yellow!40}\makebox[\linewidth]{\makecell{\textit{Comparison Study on ACDC}}}}\\
1 & 2D\&3D UNets \cite{isensee2018automatic} &91.92 &90.78 &90.46 &94.54 & 49.00 \\
2 & 2D-3D FCNN \cite{patravali20182d} &88.33 &87.00 &85.50 &92.50 & 53.00 \\
3 & TransUNet \cite{chen2021transunet}   & 89.71   & 88.96  & 84.53    & 95.73  &52.00 \\
4 & SwinUNet \cite{cao2022swin}      & 90.00  & 88.55 & 85.62  & 95.83  &48.00  \\
5 & Parallel MERIT \cite{rahman2024multi}   & 92.32   &  90.87  & 90.00   & \textbf{96.08}  &  51.00\\ 
6 & FCT  \cite{tragakis2023fully}  & \textbf{93.02}   & \textbf{92.64}  & \textbf{90.51}  & 95.90  & 48.00\\
\textbf{7} & \textbf{2UNet (Uncertainty)} &92.29  &92.11  &89.04  &95.73  &\textbf{81.00} \\
8 & 3UNet (Uncertainty) &92.16  &91.90  &88.99  &95.58  &69.00 \\
\midrule
\multicolumn{7}{c}{\cellcolor{yellow!40}\makebox[\linewidth]{\makecell{\textit{Comparison Study on M\&Ms}}}}\\
1 & UNet+DA+DUNN \cite{corral20212}    
& 86.62 & 86.30 & 83.35 & 90.20 & 76.14\\ 
2 & UNet (ResNet-34) \cite{parreno2021deidentifying}  
& 87.00 & 85.75 & 84.10 & 91.15 & 76.75 \\ 
3 & nnUNet \cite{full2021studying}     
& \textbf{88.35} & \textbf{88.50} & \textbf{85.30} & 91.25 & 79.48\\ 
4 & 2UNet (Uncertainty)  
&87.61 & 87.70 & 83.16 & 91.97 &84.93 \\ 
\textbf{5} & \textbf{3UNet (Uncertainty)}  
&88.13 & 88.36 & 83.78 & \textbf{92.25} &\textbf{88.13} \\ 
\midrule
\multicolumn{7}{c}{\cellcolor{yellow!40}\makebox[\linewidth]{\makecell{\textit{Ablation Study on ACDC}}}}\\
1 &UNet    
&91.27  &90.62  &87.93  &95.28  &53.00 \\ 
2 &D3P   
&89.45  &89.41  &85.23  &93.71  &35.00 \\ 
3 &1UNet+1D3P (\textit{Fixed})
&90.90  &90.63  &87.47  &94.60  &70.00 \\ 
4 &1UNet+1D3P (\textit{Stacking})
&91.38  &90.75  &88.07  &95.32 &54.00 \\ 
5 &1UNet+1D3P (Uncertainty)
&90.89  &90.41  &87.63  &94.62  &38.00 \\ 
6 &2UNet (\textit{Fixed})
&91.25  &91.06  &87.59  &95.10  &71.00 \\ 
7 &2UNet (\textit{Bagging})
&91.71 &90.95  &88.66 &95.53 &73.00 \\ 
8 &2UNet (\textit{Augmenting})
&69.90  &59.36  &69.60  &80.76 &22.00 \\ 
\textbf{9} &\textbf{2UNet (Uncertainty)}
&\textbf{92.29}  &\textbf{92.11}  &\textbf{89.04}  &\textbf{95.73}  &\textbf{81.00} \\ 
10 &3UNet (Uncertainty)
&92.16  &91.90  &88.99  &95.58  &69.00 \\ 
11 &4UNet (Uncertainty)
&91.98  &91.80  &88.71  &95.44  &68.00 \\ 
\bottomrule

\label{tab:results}
\end{tabular}
\end{table}

\textbf{\textit{Ablation on ACDC}}. First, we evaluated Streaming with \textit{Fixed} to identify the optimal weights as a baseline for ablation studies. Visual results are in Fig.~\ref{fig:convex comb comprison viz}. The homogeneous combination 2UNet with weights (0.5, 0.5) achieves better performance, with an average DSC of 91.25. For the heterogeneous combination 1UNet + 1D3P, the highest DSC 90.90 is achieved with weights (0.7, 0.3). Then, we compared \textit{Stacking}, \textit{Bagging}, \textit{Augmenting} and the optimal Streaming with \textit{Fixed} (all using Weighted Average pooling with \textit{Fixed}'s weights for fairness) against our Streaming with Uncertainty on ACDC, as shown in Tab.~\ref{tab:results}'s third major row. The observations are as follows: \textbf{obs 1)} Our Streaming with Uncertainty outperforms traditional ensemble method, and 2UNet with Uncertainty is the most stable on DSC and EC. We attribute this to that heterogeneous ensembles bring diversity but causes \textit{the veto effect} \cite{shazeer2017outrageously}, where the parameters of a dominant model are optimized rapidly while the parameter updates of other members stagnate, making homogeneous ensembles perform better. \textbf{obs 2)} Increasing the number of classifiers does not consistently improve segmentation performance, and two homogeneous classifiers achieve the best results. It endorse Webb et al.'s view \cite{webb2021ensemble} that both UNet and D3P are high capacity models. As the model number increases, the difference between E2E and independent training vanishes, leading to a decline in segmentation performance. 

\textit{\textbf{Case Study}}. We visualized a challenging sample from the ACDC testset in Fig.~\ref{fig:cmr viz}, the ES frame of patient114 from the hypertrophic cardiomyopathy (HCM) group. The blue box highlights the best-performing model, while the green box shows challenges in other baselines. The last two rows of the first column show \(\boldsymbol{\sigma}_{i}\) and \(\bar{\boldsymbol{\omega}}_{i}\). Uncertainty is mainly concentrated at ventricular and chamber boundaries.

\textbf{\textit{Computational Efficiency}}. Ensemble learning often faces complexity dilemma. Compared to baselines on ACDC in Fig. \ref{fig:param}, we achieve near-SOTA Average DSC with lightweight and deployment-friendly parameters and FLOPs.

\section{Conclusion}

In this work, we propose an end-to-end ensemble method, Streaming. Evaluating on ACDC and M\&Ms CMR datasets, Streaming with Uncertainty achieves near-SOTA performance on DSC and outperforms all baselines on EC, making it more suitable for patient-specific and clinical applications.

\section{Funding Acknowledgments}
We would like to thank the support from the National Natural Science Foundation of China under Grants No. 62406206.


\bibliographystyle{IEEEbib}
\bibliography{main}

\end{document}